\documentclass[conference,10pt]{IEEEtran}
\IEEEoverridecommandlockouts
\usepackage{cite}
\usepackage{amsmath,amssymb,amsfonts}
\usepackage{algorithm}
\usepackage{algorithmicx}
\usepackage{algpseudocode}
\usepackage{graphicx}
\usepackage{textcomp}
\usepackage{xcolor}
\usepackage{lettrine}
\usepackage{multirow}
\usepackage{caption}
\usepackage{subcaption}
\usepackage{tabularx}
\usepackage{romannum}
\usepackage{array}
\usepackage{cleveref}
\usepackage[textwidth=30mm]{todonotes}

\begin{document}
\title{BiLSTM and Attention-Based Modulation Classification of Realistic Wireless Signals \vspace*{-0.5em}}
\author{Rohit Udaiwal, Nayan Baishya, Yash Gupta, and B. R. Manoj \\
Department of Electronics \& Electrical Engineering, 
Indian Institute of Technology Guwahati, Assam, India\\
Emails: {\tt{\{udaiwalrohit, yashguptaji2001\}@gmail.com,  \{nmb94, manojbr\}@iitg.ac.in}}
\vspace*{-1.35em}
\normalsize
\thanks{This work was supported by SERB Start-Up Research Grant (SRG) Scheme, Department of Science and Technology (DST), Govt. of India under Grant SRG/2022/001214.}
}




\maketitle

\begin{abstract}
This work proposes a novel and efficient quad-stream BiLSTM-Attention network, abbreviated as QSLA network, for robust automatic modulation classification (AMC) of wireless signals. The proposed model exploits multiple representations of the wireless signal as inputs to the network and the feature extraction process combines convolutional and BiLSTM layers for processing the spatial and temporal features of the signal, respectively.  An attention layer is used after the BiLSTM layer to emphasize the important temporal features. The experimental results on the recent and realistic RML22 dataset demonstrate the superior performance of the proposed model with an accuracy up to around \textbf{$99\%$}. The model is compared with other benchmark models in the literature in terms of classification accuracy, computational complexity, memory usage, and training time to show the effectiveness of our proposed approach. 

\end{abstract}


\IEEEpeerreviewmaketitle
\vspace*{-2em}
\section{Introduction}
Automatic modulation classification (AMC) is the identification of the modulation scheme of a wireless signal under unknown prior information, such as channel conditions and transmitting equipment parameters. AMC is a blind technique used as an intermediate signal processing module in practical civilian and military applications, such as cognitive radio, threat assessment, and spectrum monitoring \cite{Huang_deep_learning_2020_new}. Conventionally, researchers have employed probabilistic-based decision-theory methods for AMC where accurate physical and signal models are known. Recently, usage of deep learning (DL) methods for AMC has become popular where the accurate signal models are unknown due to the complex nature of wireless networks or where DL has powerful computational advantages \cite{o2016convolutional, dual-lstm-2020, RML22, dual-coms-2023, dl-ws2018 }. DL-based techniques can learn potential features directly from raw inputs, owing to the significant availability of computational resources. This work focuses on developing an efficient and reliable DL solution for the automatic classification of wireless signals that outperforms the existing approaches \cite{dual-lstm-2020, RML22, dual-coms-2023, Liu_dl_amc_2017 }.

The remarkable success of DL algorithms for AMC can be attributed to the development of several large-scale synthetic radio-frequency (RF) signal modulation datasets in recent years. Deepsig's RML16 \cite{vtcnn2} and RML18 \cite{8267032} datasets are considered benchmarks, and many DL models for AMC are proposed in the literature using these datasets. More recently, a new RF dataset, named RML22 \cite{RML22}, is proposed to overcome the shortcomings of RML16 such as errors and ad-hoc parameter choices, by improving the quality of the dataset. RML22 in \cite{RML22} is a realistic dataset for AMC of wireless signals as it captures a wide range of channel conditions and hardware artifacts, and the order of artifacts in dataset generation is also corrected. In addition, RML22 has efficiently modeled the impacts of information sources on the data, specifically in analog modulation types. Therefore, developing accurate and efficient DL models for the realistic RML22 dataset is essential for better performance of real-life applications of AMC. 

In the literature, DL methods based on various architectures, such as convolutional neural network (CNN) \cite{o2016convolutional, Liu_dl_amc_2017}, long short-term memory (LSTM) \cite{dual-lstm-2020}, and attention mechanism \cite{dual-coms-2023} are proposed for AMC. Moreover, different forms of data sources as input to a DL model, such as the in-phase/quadrature ($\mathrm{I/Q}$) and/or the amplitude/phase ($\mathrm{A/\varphi}$) information are utilized to learn effective features \cite{dual-coms-2023, dual-lstm-2020}. In this work, we propose a DL model that utilizes four CNN-based streams, where each stream extracts spatial features from different forms of signal data, namely the $\mathrm{I/Q}$, $\mathrm{A/\varphi}$, and the independent $\mathrm{I}$ and $\mathrm{Q}$ information. The spatial features are fused and then passed through a temporal feature extraction block consisting of a bidirectional LSTM (BiLSTM) layer and an attention layer\cite{attention_dl_2023}. The attention layer re-weights the temporal features learned by the BiLSTM layer to emphasize the most informative feature values for AMC. The main contributions of this work are: (i) We propose a novel quad-stream BiLSTM-Attention (QSLA) architecture that achieves the best classification accuracy on the realistic RML22 RF signal dataset when compared to five benchmark DL models in the literature \cite{dual-lstm-2020}, \cite{RML22}, \cite{dual-coms-2023}, \cite{Liu_dl_amc_2017}; (ii) The computational complexity of QSLA is evaluated in terms of memory, the number of trainable parameters, and training time and is found to provide significant computational efficiency; (iii) In contrast to \cite{dual-lstm-2020} and \cite{dual-coms-2023}, we show that the early fusion of different forms of spatial features, followed by temporal feature extraction achieves superior accuracy than independent temporal feature extraction from different types of input.
\begin{figure*}[t!]
\centering
    \includegraphics[width = 6.0in,height=2.2in]{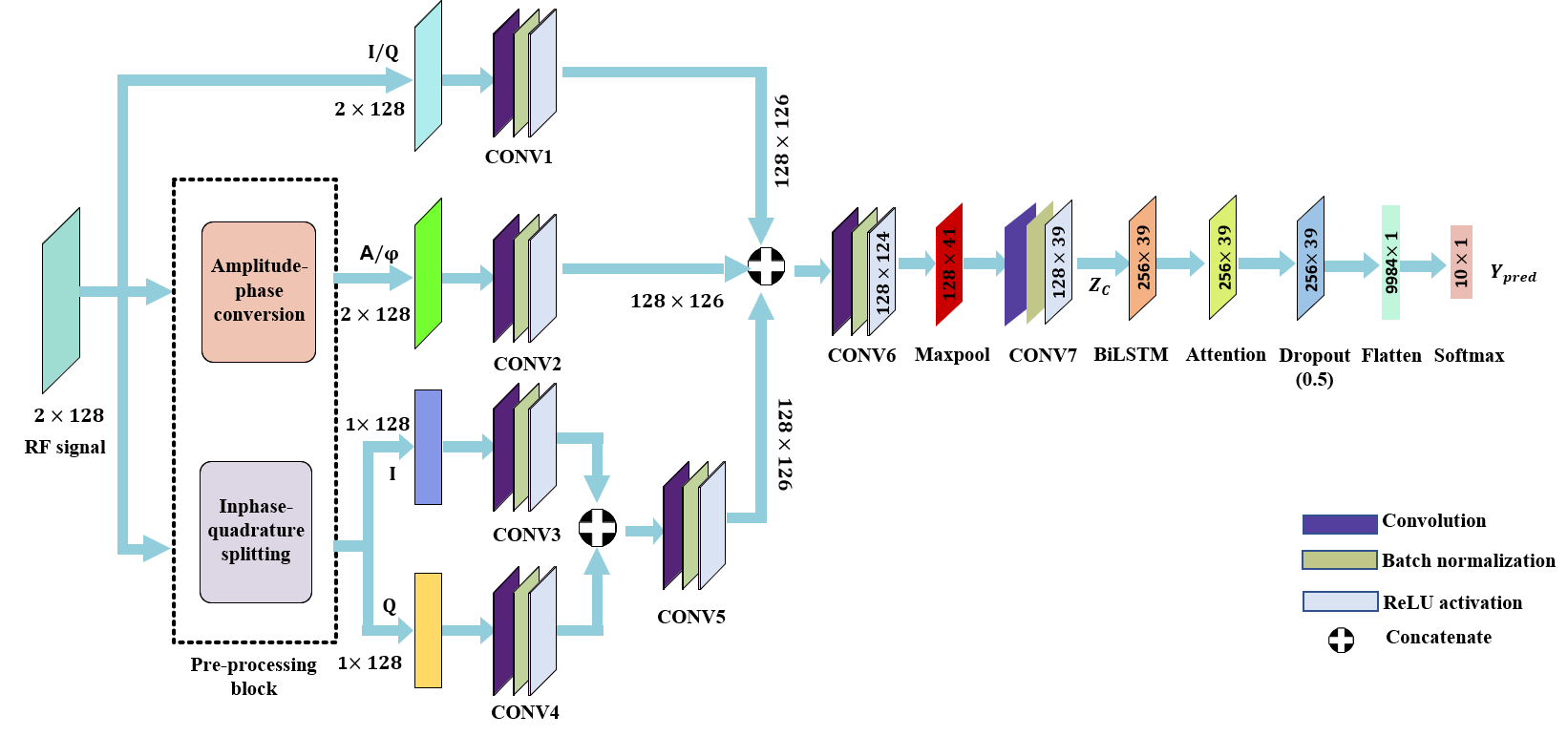}
    \vspace*{-1.25em}
    \caption{Proposed architecture: Quad-Stream BiLSTM-Attention (QSLA) network. 
    \vspace*{-2em}
    }
    \label{fig:QSLA}
\end{figure*}
\vspace*{-0.4em}
\section{Proposed Architecture}
The proposed QSLA architecture is shown in Fig. \ref{fig:QSLA}, and a description of the building blocks of the model is given below: 
\subsection{Signal Pre-processing Block}
We consider a single-input-single-output wireless system where RF signals are transmitted over orthogonal carrier waves. An RF signal $s$ as a function of time $n$ can be represented as $s[n] = s_I[n] + js_Q[n]$,
where $s[n]$ is the baseband signal, $s_I[n]$ and $s_Q[n]$ are the $\mathrm{I}$ and $\mathrm{Q}$ components of $s[n]$, respectively. The received raw $\mathrm{IQ}$ signal is first passed through a pre-processing block, as shown in Fig. \ref{fig:QSLA}. This block splits the original RF signal into four channels: $\mathrm{A}\varphi$, $\mathrm{IQ}$, $\mathrm{I}$, and $\mathrm{Q}$. The amplitude $s_A$ and phase $s_\varphi$ information in terms of $\mathrm{I}$ and $\mathrm{Q}$ components are, respectively, given by 
\begin{equation} \label{eq2}
\small
s_A  = \sqrt{s_{I}^2 + s_{Q}^2}\, \quad \text{and}\quad
s_{\varphi} = \arctan \left(\dfrac{s_Q}{s_I}\right)\, .
\end{equation}
The independent $\mathrm{I}$ and $\mathrm{Q}$ components can provide complementary information and are useful since some potential features are lost due to amplitude and phase imbalance. 
\subsection{Spatial Feature Extraction}
The four signal representations transformed using the pre-processing block, i.e., $\mathrm{A}\varphi$, $\mathrm{IQ}$, $\mathrm{I}$, and $\mathrm{Q}$, are then fed as input to each independent stream to extract spatial features. For this purpose, 1-D convolutional layers (CONV 1D) are used as they reduce the computational complexity. The CONV 1D layer in each stream consists of $128$ filters of kernel size $3$, followed by a batch normalization (BN) layer and a $\mathrm{ReLU}$ activation layer. These are represented as CONV1, CONV2, CONV3, and CONV4 in Fig. \ref{fig:QSLA}. The spatial features obtained from the $\mathrm{I}$ and $\mathrm{Q}$ components at the outputs of $\mathrm{CONV3}$ and $\mathrm{CONV4}$ are concatenated, followed by a CONV 1D layer ($\mathrm{CONV5}$) with $128$ filters and kernel size of $1$. The use of kernel size $1$ helps in the point-wise combination of the uncorrelated features from the orthogonal ${\mathrm{I}}$ and ${\mathrm{Q}}$ components. Finally, the outputs of  $\mathrm{CONV1}$, $\mathrm{CONV2}$, and $\mathrm{CONV5}$ are concatenated to obtain the fused spatial features. Two additional convolutional layers, namely $\mathrm{CONV6}$ and $\mathrm{CONV7}$, capture high-level feature representations from the fused features. Both layers consist of $128$ filters of kernel size $3$. Max-pooling with a stride of $3$ is performed at the output of $\mathrm{CONV6}$ to downsample the features. BN and $\mathrm{ReLU}$ activation is used for all the convolutional layers. The final features obtained from  $\mathrm{CONV7}$, represented as $Z_c$ are further processed to capture temporal features. The primary motivation for utilizing the independent $\mathrm{I}$ and $\mathrm{Q}$ streams is that the spatial features extracted from these two orthogonal signals will inherently be uncorrelated. A DL model generally performs well when presented with different representations of the input data, as shown in \cite{dual-lstm-2020} and \cite{dual-coms-2023}. However, a learned combination of the uncorrelated features extracted from the $\mathrm{I}$ and $\mathrm{Q}$ signals will be complementary to the features learned from the $\mathrm{IQ}$ signal and, therefore, can perform better than the dual-stream models \cite{dual-coms-2023, dual-lstm-2020} as presented in Section IV.
\vspace*{-0.5em}
\subsection{Temporal Feature Extraction}
The temporal feature extraction block consists of a BiLSTM layer and an attention layer, which are described below: 
\subsubsection{BiLSTM} 
To capture the temporal information, $Z_c$ is first fed into a BiLSTM layer. An LSTM network consists of a set of memory cells and gates that regulate the flow of information. Specifically, for each memory cell, an input gate controls the information to be added to the cell state, a forget gate controls the information to be forgotten from the cell state, and an output gate determines the relevant information from the cell state to be forwarded to the subsequent cell (hidden state). It enables the model to retain essential temporal information and helps to learn long-range dependencies. The function of the LSTM layer on the features $Z_c$ can be defined as
\begin{equation} \label{eq4}
\small
\begin{split}
H_t, C_t = \mathrm{LSTM}(Z_c, H_{t-1}, C_{t-1})\, ,
\end{split}
\end{equation}
where $H_t$ and $C_t$ denote the hidden state and cell state at time $t$, respectively; $H_{t-1}$ and $C_{t-1}$ represent the previous hidden state and cell state, respectively. A BiLSTM layer is just two stacked LSTM layers, with one layer processing the input sequence in the forward direction and the other processing in the backward direction. The output at a time $t$ combines both layers' hidden states at $t$. The BiLSTM layer in our architecture consists of $128$ cells and $\mathrm{tanh}$ activation function. We use BiLSTM to allow for feature extraction by considering both future and past context information.

\subsubsection{Attention} The output of the BiLSTM layer is then processed by an attention layer. The basic idea behind attention is to allow the model to focus on specific input features when making predictions. An attention layer calculates the attention weights for each sequence element (token) \cite{attention_dl_2023}. Based on the attention weights, task-specific decision-making is performed by emphasizing the most informative features. Also, attention layers can be computed parallelly, making them computationally efficient for parallel computing hardware like graphics processing units (GPUs) and tensor processing units (TPUs). In the proposed architecture,  dot-product attention is used to calculate the attention weights of the temporal features from the BiLSTM layer and is given by
\vspace{-1pt}
\begin{equation} \label{eq4}
\small
\begin{split}
\alpha_{t} = \mathrm{softmax}(W_a \cdot H_t + b_a)\, , 
\end{split}
\vspace{-3pt}
\end{equation}
where $\alpha_{t}$ represents the attention weights of a hidden state $H_t$,  $W_a$ and  $b_a$ are the learnable weights and biases of the attention layer, respectively, and  $\mathrm{softmax}$ is the activation function. A dropout layer with $p=0.5$ is used on top of the attention layer as a regularization technique. The attention layer provides the most important features, which helps in achieving better classification accuracy with reduced complexity.
\vspace{-0.5em} 
\subsection{Fully Connected Layer and Softmax Classifier}
The output of the attention layer is flattened and passed through a $\mathrm{softmax}$ classification layer with ten output classes to obtain the predicted probabilities for each class as given by,
\begin{equation} \label{eq4}
\small
\begin{split}
Y_{pred} = \mathrm{softmax}(W_f \cdot H_{att} + b_f)\, ,
\end{split}
\end{equation}
where $Y_{pred}$ represents the predicted class probabilities, $W_{f}$ and $b_{f}$ denote the weights and biases of the fully connected layer, respectively, and $H_{att}$ is the output of the attention layer. Further, ${\mathcal{L}}_2$ regularization is used on the weights of this layer to avoid overfitting. Thus, the proposed model implements an early fusion stage of spatial features and then extracts temporal dependencies within the features, enabling high classification performance while reducing computational complexity.
\vspace{-0.25em}
\section{Experiments and results}
\subsection{Experimental Setup}
To enable reproducibility, we have considered the publicly available RML22 \cite{RML22} modulation classification dataset. Since the realistic RML22 dataset overcomes the shortcomings of the previous RML2016.10A dataset, it is important to benchmark the modulation classification performance of DL-based AMC methods on this particular dataset for real-world signal analysis. The dataset consists of RF signals generated using $10$ modulation schemes, and for each modulation scheme, the samples are generated using SNRs from $-20$ dB to $20$ dB with a step of $2$ dB. The dataset has $420,000$ complex-valued RF signals of dimension $2 \times 128$. The dataset is split into train, test, and validation data in the ratio of $8:1:1$. We use categorical cross-entropy as the loss function and the Adam optimizer is used to train all the models. The initial learning rate is set to be $0.001$ and is multiplied by a factor of  $0.4$ if the validation loss does not decrease within $8$ epochs. The models are trained for $100$ epochs using a batch size of $1024$. Early-stopping is applied to stop the training process when the validation loss is not decreasing any further. All the experiments are performed using an ${\mathrm{NVIDIA \,\,A6000\,\, GPU}}$ with 48GB DRAM and the Tensorflow-based Keras framework to implement the models.

\subsection{Results and Discussions}
We first investigate the classification performance of the proposed QSLA model by comparing it with five benchmark architectures in \cite{dual-lstm-2020, RML22},\cite{dual-coms-2023},\cite{Liu_dl_amc_2017}. We have also compared the performance of the QSLA model for each modulation scheme across all the SNRs using precision-recall (PR) curves and analyzed the variability between high and low SNRs.
In wireless communication applications, the computational cost of the DL model is critical for real-world deployment on resource-constrained edge devices, such as IoT systems and smartphones. Therefore, computational complexity measures, such as the number of trainable parameters, the memory of the saved weights, and the training time per epoch, are also used to compare all the benchmark models with the proposed model. Finally, we provide a performance analysis of different combinations of BiLSTM and attention layers in the temporal feature extraction block to emphasize the effectiveness of BiLSTM followed by attention in our network both in terms of classification accuracy and computational complexity.

                 

\begin{figure}[!t]
\centering
\includegraphics[width=2.45in,height=1.7in]{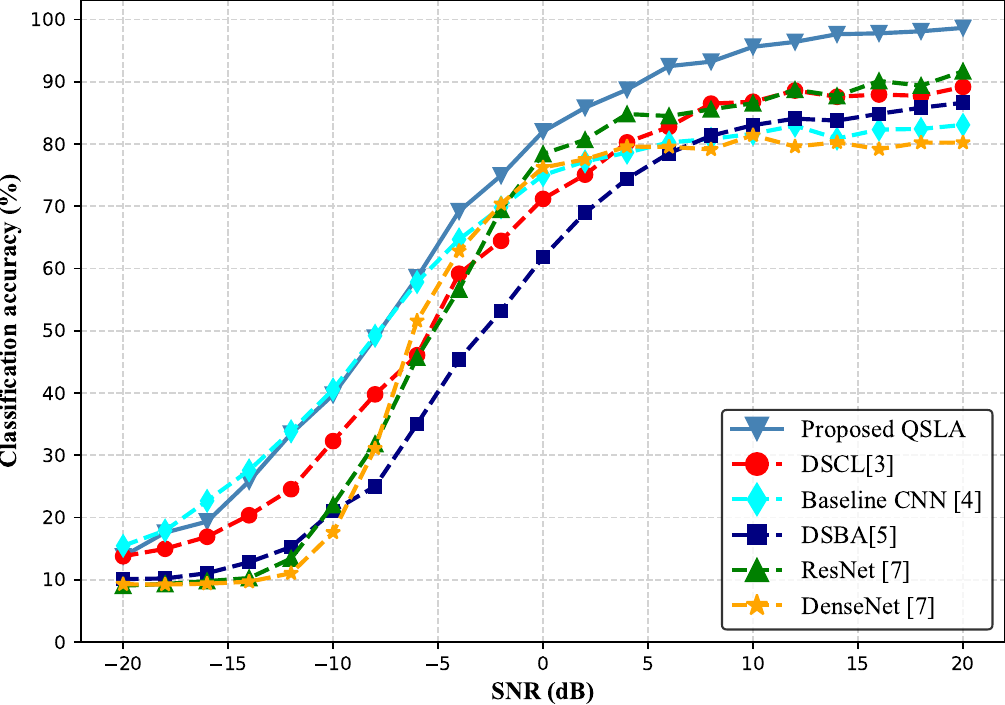}
\caption{Accuracy comparison of QSLA vs benchmark models.}
    \label{fig:all models}
\end{figure}
\subsubsection{Classification Performance}
The classification performance of the proposed model is compared to that of the five benchmark DL models considered in this work and is shown in Fig. \ref{fig:all models}. It can be observed that the accuracy of the proposed QSLA model is consistently higher than $90\%$ for SNR $\geq 4$ dB and achieves $\sim 99\%$ at high SNRs. This performance is significantly better than the baseline CNN proposed by the authors in \cite{RML22} specifically for the RML22 dataset and can achieve only up to $\sim 83\%$ even at high SNRs. However, the baseline CNN in \cite{RML22} is elementary in terms of model architecture, and a fair comparison with larger, more complex networks is required further to emphasize the better performance of the proposed QSLA model. To the best of our knowledge, there are no recent DL architectures apart from the baseline CNN that have been developed specifically for the RML22 dataset. Therefore, in this work, we have chosen four benchmark DL models presented in \cite{dual-lstm-2020, dual-coms-2023, Liu_dl_amc_2017}, which are originally developed for the RML2016 dataset and have complex architectures. We have implemented each of these architectures and then performed the training and testing with the same data splits of the RML22 dataset used to develop our QSLA network. The dual-stream CNN-LSTM (DSCL) model in \cite{dual-lstm-2020} and the dual-stream BiLSTM-Attention (DSBA) model in \cite{dual-coms-2023} are related to our model, as both the methods employ LSTM-based temporal features extraction using $\mathrm{IQ}$ and $\mathrm{A\varphi}$ representations in the two different streams. 
However, the key advantages of the proposed model architecture compared to \cite{dual-lstm-2020, dual-coms-2023} are: (a) The addition of the independent $\mathrm{I}$ and $\mathrm{Q}$ streams at the input of the model to extract uncorrelated spatial features; (b) The DSCL model \cite{dual-lstm-2020} incorporates LSTM layers separately at each stream and uses the outer product to fuse the feature maps of the two streams before performing classification. The outer product on the high-dimensional features increases the computational complexity of the model significantly. Moreover, both \cite{dual-lstm-2020} and \cite{dual-coms-2023} perform a late fusion of the temporal features extracted separately from both streams. This late fusion of temporal features may not be computationally efficient when scaling to multiple streams of input as the cost of the LSTM and attention layers are generally higher. Therefore, the proposed model uses an early fusion of spatial features extracted from various signal representations without using the outer product, and then temporal features are extracted using only a single BiLSTM and attention layers. Apart from these, the two other complex CNNs used for comparison are the ResNet and the DenseNet \cite{Liu_dl_amc_2017}.

As with the baseline CNN, the comparison of the accuracy of our QSLA model with all the four complex architectures in Fig. \ref{fig:all models} shows that the performance of the QSLA model is significantly higher.
The DSCL model can only achieve up to $\sim 90\%$ accuracy at high SNRs for the RML22 dataset. Similarly, the performance is better than DSBA, ResNet, and DenseNet architectures. At SNR$=16$dB, the accuracy of the DSCL, baseline CNN, DSBA, ResNet, and DenseNet models are $87.9\%$, $82.35\%$, $84.8\%$, $90.1\%$, and $79.1\%$, respectively; whereas the proposed QSLA model achieves an accuracy of $97.8\%$. Thus, our proposed model outperforms all the architectures by a significant margin. 
\begin{figure}[!t]
\centering
\includegraphics[width=2.45in,height=1.7in]{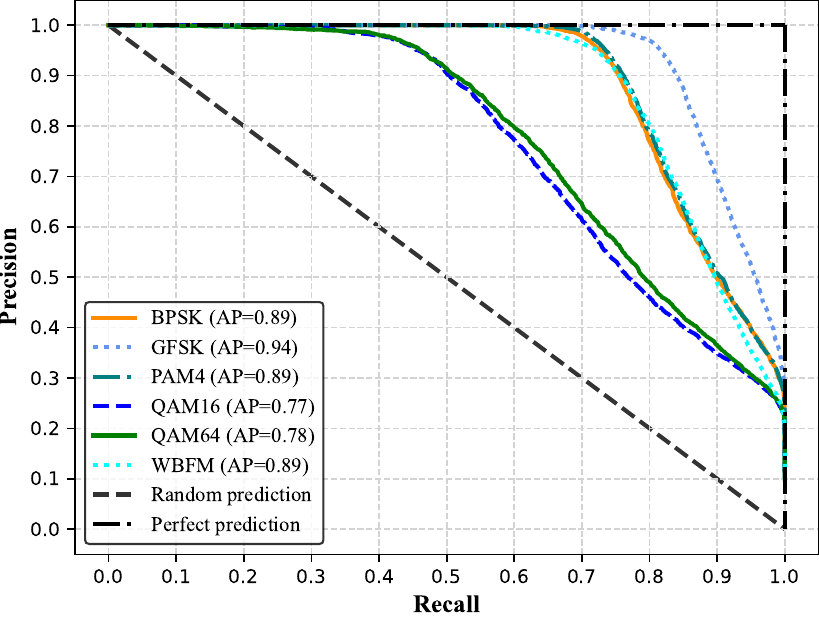}
\caption{Classwise PR curves for QSLA model across all SNRs.}
    \label{fig:all-pr-re}
\end{figure}
                 
\begin{figure}[!]
\centering
        \subfloat[]{\includegraphics[width=2.15in,height=1.7in]{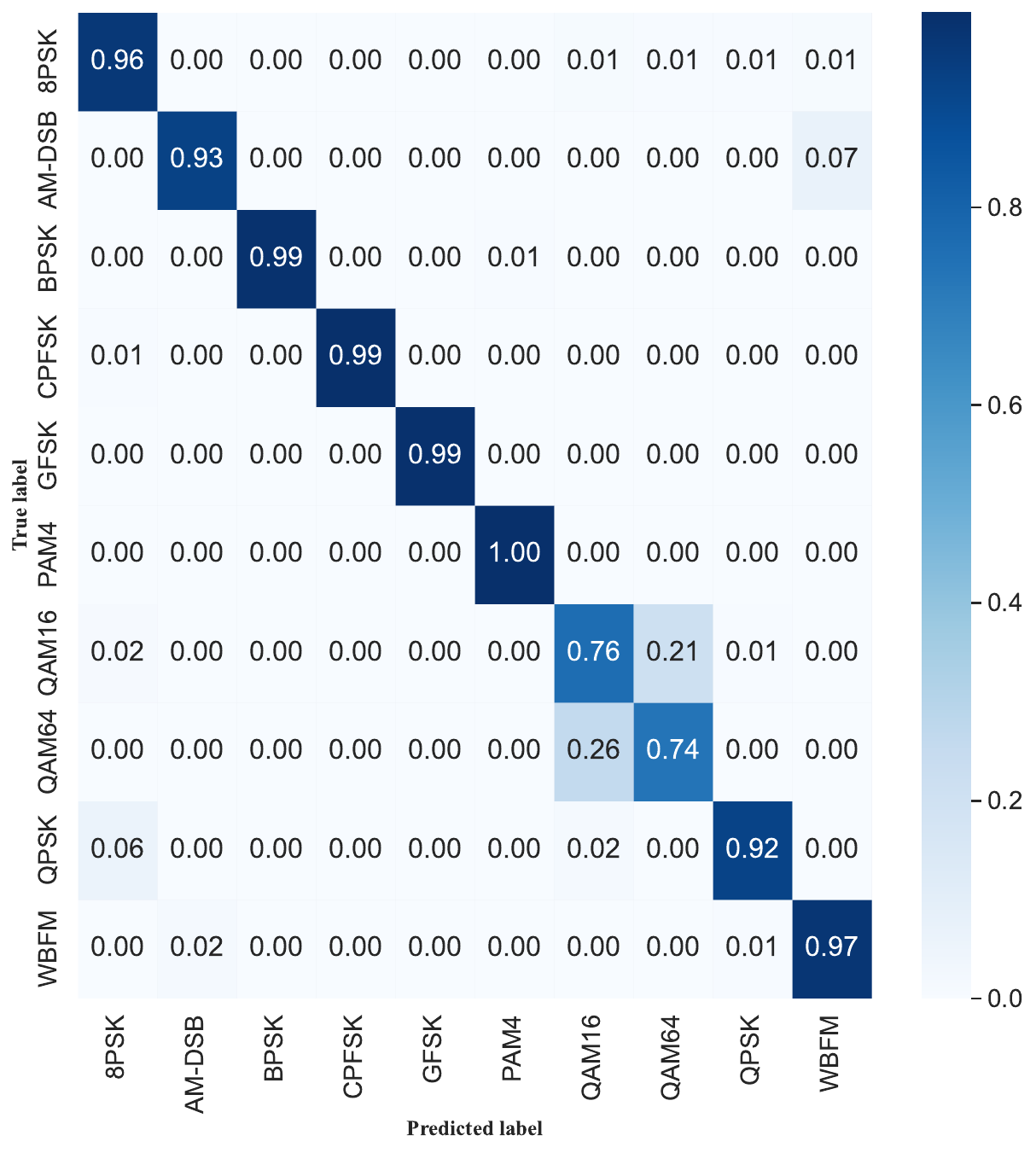}%
                 \vspace*{-0.25em}
                 \label{fig: qsla_cm_6db}}
                 
        \subfloat[]{\includegraphics[width=2.15in,height=1.7in]{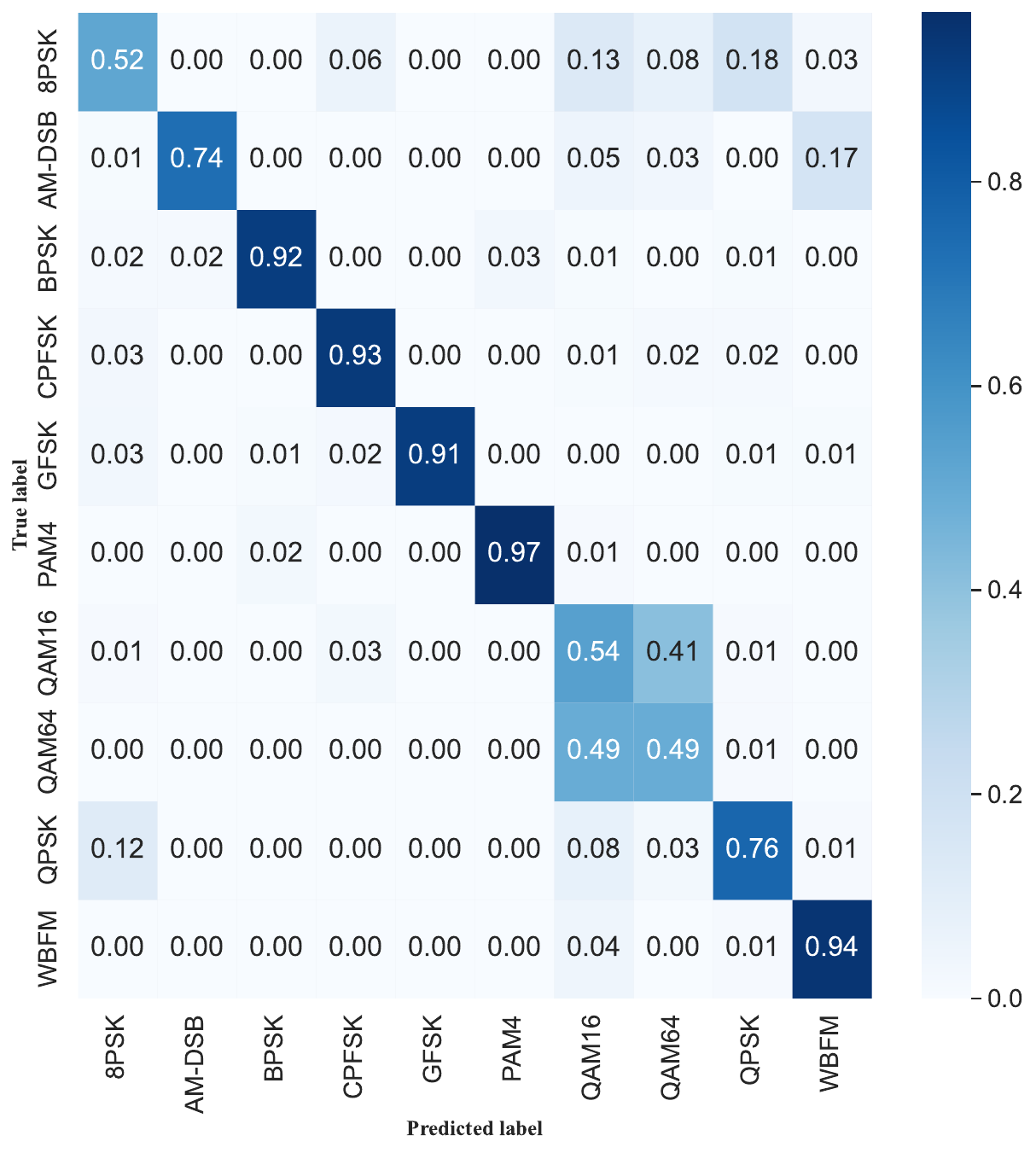}%
                \vspace*{-0.25em}
                \label{fig: dsba_cm_6db}}
        \caption{Confusion matrices at $6$dB: (a) Proposed QSLA model and (b) DSBA model \cite{dual-coms-2023}.}
        \label{fig:cm_comp_6db}
\end{figure}
We have also analyzed the performance of all the modulation schemes individually using PR curves.
The precision and recall values for the class are obtained at varying thresholds of the model, which determines the decision boundary between predicting positive and negative instances. 
The average precision ($AP$) can be defined as the mean precision score across various thresholds on the PR curve, where the weight assigned to each score corresponds to the change in recall from the previous threshold. Fig. \ref{fig:all-pr-re} compares the PR curves of different modulation classes using the predictions for all SNRs and reports the corresponding $AP$ values. It can be observed that the QAM16 and QAM64 modulations perform the least among all, with  $AP$ values of $0.77$ and $0.78$, respectively. We have also observed that the performance for a particular modulation may vary with SNR. For example, at SNR$=18$dB, all the modulations have $AP$ values close to $1$, depicting near-perfect prediction performance. However, at SNR$=-6$dB, QAM16 and QAM64 have $AP$ values of $0.48$ and $0.55$, respectively, which is significantly lower than that of other modulation schemes. This shows that the less performance of QAM16 and QAM64 as shown in Fig. \ref{fig:all-pr-re} is mainly contributed by the inaccurate predictions made at the low SNRs. The overlapping constellation points between the two schemes can be a reason for the lower precision for these modulations. This is further impacted by the fact that at low SNRs, the noise power is significant compared to the signal power, which affects the feature extraction from the RF signals at these SNRs. 

Furthermore, we have compared the performances of the benchmark models at a low SNR with the proposed QSLA model. Fig. \ref{fig: qsla_cm_6db} and Fig. \ref{fig: dsba_cm_6db} show the confusion matrices of the QSLA and the DSBA \cite{dual-coms-2023} models at SNR$=6$dB, respectively. It can be observed that for multiple modulation schemes, i.e., 8PSK, AM-DSB, QAM16, QAM64, and QPSK, the DSBA model has much less true positive (TP) ratios than the QSLA model. Specifically, the QSLA model has a TP ratio of $0.96$ for 8PSK compared to $0.52$ for the DSBA model. For QAM16 and QAM64, the DSBA model can only achieve TP ratios of $0.54$ and $0.49$, respectively. Though the values are considerably higher for the QSLA model, i.e., $0.76$ for QAM16 and $0.74$ for QAM64, there is still scope for improvement when compared to other modulations. This is an aspect of DL model development for AMC that needs to be emphasized for more reliable signal recognition at practical low SNR scenarios.


\subsubsection{Computational Complexity}
\begin{figure}[!]
\centering
\includegraphics[width=2.15in,height=1.5in]{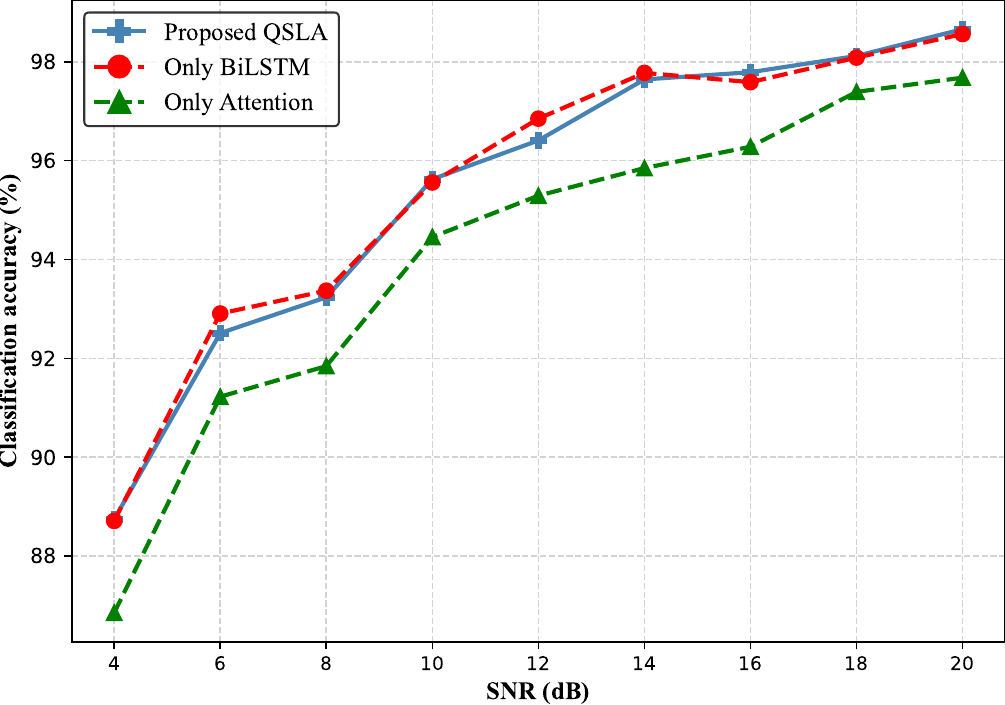}
\caption{Performance of the proposed model with different layers for temporal features extraction.}
    \label{fig:layers}
\end{figure}

The key objective of our work is to achieve 
excellent classification performance with computational advantage. Therefore, we compare the computational complexities of the proposed and the benchmark architectures considered in this work in Table \ref{table:1}. It can be observed that compared to the models in \cite{dual-lstm-2020} and \cite{Liu_dl_amc_2017}, our model has significantly less complexity in terms of memory, trainable parameters, and training time per epoch.
The DSBA model in \cite{dual-coms-2023} has fewer parameters than our proposed model, which can be attributed to the fact that the DSBA model was originally developed for the classification of only six modulation schemes.
As mentioned in the previous discussion, the authors in \cite{dual-lstm-2020} have used LSTM layers in each of the two streams of the DSCL model for temporal feature extraction, which increases the complexity of the network. We have shown that a single BiLSTM layer can extract the temporal features, which reduces the number of trainable parameters and the convergence time of the model. We have achieved almost $35\%$ reduction in trainable parameters compared to \cite{dual-lstm-2020} while significantly improving classification performance on RML22.
\begin{table}[!t]
\centering
\caption{Comparison of computational complexities of the proposed QSLA architecture with other benchmark architectures\vspace*{-0.3em}}\label{table:1}
\renewcommand{\arraystretch}{1.3}
\resizebox{\columnwidth}{!}{ 
    \begin{tabular}{| c | c | c | c |}
    \hline
     &  Memory ($\mathrm{kB}$) & Trainable parameters ($\mathrm{k}$) & Training time ($\mathrm{s}$)/epoch \\ \hline
    Baseline CNN \cite{RML22} & $345$ & $118$ &  $27$\\ \hline
    DSBA \cite{dual-coms-2023} & $610$ & $144$ &  $62$\\ \hline
    {\textbf{Proposed QSLA}} & $2,502$ & $615$ &  $96$\\ \hline
    DSCL \cite{dual-lstm-2020} & $4,532$ & $1,142$ &  $190$\\ \hline
    ResNet \cite{Liu_dl_amc_2017} & $11,800$ & $3,098$ & $323$\\ \hline
    DenseNet \cite{Liu_dl_amc_2017} & $12,500$ & $3,282$ & $337$ \\ \hline
    \end{tabular}
}
\end{table}
Furthermore, we have compared the proposed model's performance with two other alternative combinations of BiLSTM and attention layers: in the model named only BiLSTM, the attention layer is replaced with another BiLSTM layer, and in the model named only attention, the BiLSTM layer is replaced with an additional attention layer. Fig. \ref{fig:layers} compares the classification performances and Table \ref{table:2} compares the computational complexities of all three combinations. It can be observed from Fig. \ref{fig:layers} that both the QSLA model and the only BiLSTM model have comparable classification performances, but using an attention layer in the QSLA model instead of an additional BiLSTM layer reduces the parameters by about $378,000$ as seen in Table \ref{table:2}, i.e., almost $38\%$ decrease in trainable parameters and hence a significant decrease in the training time. 
On the other hand, the performance of the model using only attention layers is slightly lower than the proposed QSLA model, though it has about $50\%$ fewer parameters. Thus, it can be concluded that a combination of BiLSTM and attention layers performs the best when both classification accuracy and computational complexity are considered together, compared to different benchmark models and the model using only BiLSTM layers. If the reduction in computational complexity is the primary goal, then we can consider the model using only attention layers with some compromise in accuracy; this can serve as an alternate model with an even faster convergence time than the proposed QSLA model.

\begin{table}[!t]
\centering
\caption{Comparison of computational complexities of different layers in the proposed QSLA architecture \vspace*{-0.5em}}\label{table:2}
\renewcommand{\arraystretch}{1.2}
\resizebox{\columnwidth}{!}{
    \begin{tabular}{| c | c | c | c |}
    \hline
     &  Memory ($\mathrm{kB}$) & Trainable parameters ($\mathrm{k}$) & Training time ($\mathrm{s}$)/epoch \\ \hline
      Proposed QSLA & $2,502$ & $615$ &  $96$ \\
      \hline
     Only Attention & $1,274$ & $302$ &  $45$ \\ \hline
    Only BiLSTM & $3,982$ & $993$ &  $170$ \\ \hline
    \end{tabular}  
     
}
\end{table}
\vspace{-0.5em} 
\section{Conclusions}
In this work, we have proposed a quad-stream BiLSTM-Attention model (QSLA) for the AMC task of RF signals. In the model, spatial features are extracted from $\mathrm{IQ}$, $\mathrm{A}\varphi$, $\mathrm{I}$, and $\mathrm{Q}$ representations of the input RF signal. These features are fused and then processed by a single temporal feature extraction module consisting of a BiLSTM and an attention layer. We have evaluated our method on the realistic RML22 dataset, achieving a state-of-the-art classification accuracy of about $99\%$ at high SNRs. The experiments comparing the proposed and other benchmark models from the literature show a remarkable improvement in the performance of our proposed model in terms of classification accuracy, trainable parameters, and memory footprint. We have also elaborated on using attention as an alternative to conventional BiLSTM to perform efficient temporal feature extraction while reducing model complexity. Future work will involve developing DL models that perform better at low SNRs and are robust against security threats.

\bibliographystyle{IEEEtran}

\bibliography{RG_NMB_YG_MBR_SPCOM24}

\begin{thebibliography}{10}
\providecommand{\url}[1]{#1}
\csname url@samestyle\endcsname
\providecommand{\newblock}{\relax}
\providecommand{\bibinfo}[2]{#2}
\providecommand{\BIBentrySTDinterwordspacing}{\spaceskip=0pt\relax}
\providecommand{\BIBentryALTinterwordstretchfactor}{4}
\providecommand{\BIBentryALTinterwordspacing}{\spaceskip=\fontdimen2\font plus
\BIBentryALTinterwordstretchfactor\fontdimen3\font minus \fontdimen4\font\relax}
\providecommand{\BIBforeignlanguage}[2]{{%
\expandafter\ifx\csname l@#1\endcsname\relax
\typeout{** WARNING: IEEEtran.bst: No hyphenation pattern has been}%
\typeout{** loaded for the language `#1'. Using the pattern for}%
\typeout{** the default language instead.}%
\else
\language=\csname l@#1\endcsname
\fi
#2}}
\providecommand{\BIBdecl}{\relax}
\BIBdecl

\bibitem{Huang_deep_learning_2020_new}
H.~{Huang} \emph{et~al.}, ``Deep learning for physical-layer 5{G} wireless techniques: Opportunities, challenges and solutions,'' \emph{IEEE Wireless Commun.}, vol.~27, no.~1, pp. 214--222, Feb. 2020.

\bibitem{o2016convolutional}
T.~J. O’Shea, J.~Corgan, and T.~C. Clancy, ``Convolutional radio modulation recognition networks,'' in \emph{Proc. Int. Conf. Eng. Appl. Neural Netw. (EANN)}, Aberdeen, UK, Sep. 2-5, 2016, pp. 213--226.

\bibitem{dual-lstm-2020}
Z.~Zhang, H.~Luo, C.~Wang, C.~Gan, and Y.~Xiang, ``Automatic modulation classification using {CNN-LSTM} based dual-stream structure,'' \emph{IEEE Trans. Veh. Technol.}, vol.~69, no.~11, pp. 13\,521--13\,531, Nov. 2020.

\bibitem{RML22}
V.~Sathyanarayanan, P.~Gerstoft, and A.~E. Gamal, ``{RML22}: Realistic dataset generation for wireless modulation classification,'' \emph{IEEE Trans. Wireless Commun.}, vol.~22, no.~11, pp. 7663--7675, Nov. 2023.

\bibitem{dual-coms-2023}
A.~Parmar, K.~A. Divya, A.~Chouhan, and K.~Captain, ``Dual-stream {CNN-BiLSTM} model with attention layer for automatic modulation classification,'' in \emph{Proc. 15th Int. Conf. on Commun. Syst. {\&} Netw. (COMSNETS)}, Bangalore, India, Jan. 3-8, 2023, pp. 603--608.

\bibitem{dl-ws2018}
S.~Rajendran \emph{et~al.}, ``Deep learning models for wireless signal classification with distributed low-cost spectrum sensors,'' \emph{IEEE Trans. Cognitive Commun. and Netw.}, vol.~4, no.~3, pp. 433--445, Sep. 2018.

\bibitem{Liu_dl_amc_2017}
X.~Liu, D.~Yang, and A.~E. Gamal, ``Deep neural network architectures for modulation classification,'' in \emph{Proc. 51st Asilomar Conf. on Signals, Syst., and Comput.}, Pacific Grove, CA, USA, Oct. 29-Nov. 01, 2017, pp. 915--919.

\bibitem{vtcnn2}
T.~O'Shea and N.~West, ``Radio machine learning dataset generation with {GNU} {R}adio,'' in \emph{Proc. GNU Radio Conf.}, Boulder, CO, USA, Sep. 2016.

\bibitem{8267032}
T.~J. O’Shea, T.~Roy, and T.~C. Clancy, ``Over-the-air deep learning based radio signal classification,'' \emph{IEEE J. Sel. Topics in Sig. Process.}, vol.~12, no.~1, pp. 168--179, Feb. 2018.

\bibitem{attention_dl_2023}
G.~Brauwers and F.~Frasincar, ``A general survey on attention mechanisms in deep learning,'' \emph{IEEE Trans. Knowl. and Data Eng.}, vol.~35, no.~4, pp. 3279--3298, 1 Apr. 2023.

\end{thebibliography}
 
\end{document}